# Reinforced Disentanglement for Face Swapping without Skip Connection

Xiaohang Ren, Xingyu Chen, Pengfei Yao, Heung-Yeung Shum, Baoyuan Wang

Xiaobing.AI

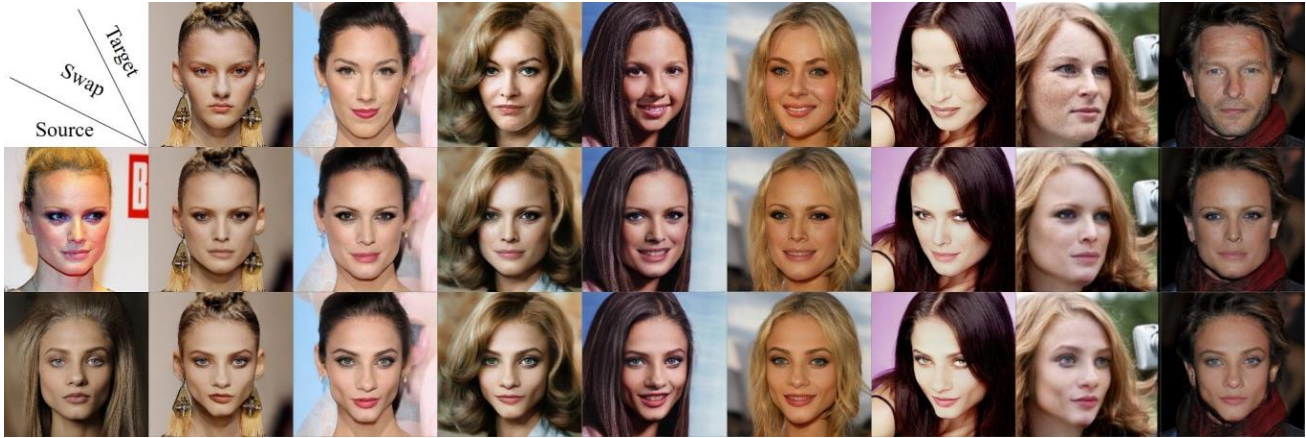

Figure 1. Face swap results by our method. Note that, the swapped images consistently preserve the identity of the source and the non-identity attributes(i.e., hair, expression, gaze, pose,.etc.) from a variety of target images. Best viewed with zoom-in and a colored version.

## Abstract

*The SOTA face swap models still suffer the problem of either target identity (i.e., shape) being leaked or the target non-identity attributes (i.e., background, hair) failing to be fully preserved in the final results. We show that this insufficient disentanglement is caused by two flawed designs that were commonly adopted in prior models: (1) counting on only one compressed encoder to represent both the semantic-level non-identity facial attributes(i.e., pose) and the pixel-level non-facial region details, which is contradictory to satisfy at the same time; (2) highly relying on long skip-connections [50] between the encoder and the final generator, leaking a certain amount of target face identity into the result. To fix them, we introduce a new face swap framework called "WSC-swap" that gets rid of skip connections and uses two target encoders to respectively capture the pixel-level non-facial region attributes and the semantic non-identity attributes in the face region. To further reinforce the disentanglement learning for the target encoder, we employ both identity removal loss via adversarial training (i.e., GAN [18]) and the non-identity preservation loss via prior 3DMM models like [11]. Extensive experiments on both FaceForensics++ and CelebA-HQ show that our results significantly outperform previous works on a rich set of metrics, including one novel metric for measuring identity consistency that was completely neglected before.*

## 1. Introduction

Deepfake [27] is a double-edged sword, despite the fact that various negative use cases are currently dominating the conversations to steer people's attention to their social impact [21], it's arguably true that AI-synthesized faces, bodies, and voices have huge potential for a variety of positive applications, such as virtual instructor or health counseling as discussed in [48], or other content generations in movie/game industry (i.e., Paul Walker in "Fast and Furious 7") and etc. Keep improving the generation quality also inspires new ways to detect [7, 15, 20, 57] deepfakes for negative use cases. Face swapping has been studied intensively in both academia and industry, where the quality improvement is remarkable over the years thanks to the advances in deep generative learning. As defined in [32], face swapping must achieve three goals simultaneously (1) fully preserve the face identity from the source image (2) fully preserve everything else except the identity (identity-irrelevant) from the target image, and (3) ensure the final result is both artifacts-free and photo-realistic.

Besides, in real-world scenarios, a source image could be swapped to multiple targets, where cross-target consistency becomes fairly important. Therefore, we additionally argue that the disentanglement of ID and non-ID properties is essential for cross-target consistent face swapping. Prior works [8, 17, 32, 35, 58] investigated various methodologies for face identity and non-identity attributes disentanglement as well as their fusions to synthesize the final swapped image. Despite the impressive progress, the above three goals are hardly achieved at the same time. One of the key challenges is that those prior works are essentially playing the seesaw-style game, where improving on the identity preservation from the source is usually at the cost of sacrificing the non-identity preservation from the target, and vice versus. Fundamentally, we argue that this is still caused by the insufficient disentanglement between face identity and non-identity representations. Specifically, on one hand, works that try to improve the source identity preservation through 3D prior [58] or information compression [17] or smoothing regularization [32] on top of a pre-trained face recognition model [35] usually don't provide sufficient forces to remove the target identity while preserving other non-identity attributes. On the other hand, work like Faceshifter [35] that tries to preserve the low-level non-identity details through attribute loss would also likely leak target identity information into the results. Such phenomena are evident from Fig. 3, where the target face silhouette is clearly leaked to the final swapped faces of works [8, 35].

By delving deep into the model structures, we noticed that most of the prior works [8, 17, 32, 35, 58] leverage a bottleneck encoder and decoder structure on the target image, where the encoder is expected to (1) fully remove the face identity (2) fully preserve everything else (i.e., pixel-level background appearance, hair, facial expression, head pose, eye gaze, and other non-identity facial attributes) from the target image. While the decoder (also called the generator) is in charge of generating the final swapped result by fusing the source identity representation from a face recognition model as well as all the compressed representations from the target decoder. To restore more details, symmetrical long skip connections are often employed by copying fine-grained low-level features from the encoder [8, 17, 35]. However, we argue that it is very difficult, if not impossible, to simultaneously achieve the above two goals using only one single compressed bottleneck encoder(see $\mathbf{Z}^{nid}$ in Fig. 2 (a)). In addition, the skip connections used in the decoder would inevitably bring the target identity information into the results together with other non-identity attributes, therefore, further hurting the disentanglement learning.

To encourage thorough disentanglement, in this paper, we designed a new framework that gets rid of the skip connections in our entire model structure. Specifically, we proposed two separate encoders that respectively capture the pixel-level attributes outside the face region and the semantic-level non-identity facial attributes inside the face region. Each of which is tailored to dedicated capturing its own desired representation without compromising with each other. Meanwhile, a target identity removal loss and a few non-identity attribute preservation losses are explicitly employed to compensate for the missing details due to the lack of skip connections. To be specific, we leverage adversarial training techniques to penalize the target representation if it contains any target identity, meanwhile, we use prior models like 3DMM predictor [14] to explicitly preserve facial expression and head pose, in addition to the attribute loss proposed in Faceshifter [35].

To summarize, our contributions are (1) we analyzed that prior works are still suffering poor disentanglement where improving one goal may hurt other goals; (2) we unveiled that the skip connection is one root cause for such insufficient disentanglement in prior model architecture; (3) we proposed new network structures, as well as new training strategies to reinforce the identity disentanglement while preserving more identity-irrelevant attributes to compensate the lack of skip connections; lastly (4) we introduced a new evaluation metric and conducted extensive experiments, the results validated the effectiveness of our method.

## 2. Related works

**Face Swap**. Early face swap works [3, 10, 37, 44] mainly leverage the 3D facial model to transfer facial identity from the source into target images/videos. Recent works heavily rely on Generative Adversarial Networks (GAN) to improve the face swap visual quality. Korshunova et al. [33] design a person-specific network for face swapping, in which each source identity model is trained separately. Deepfakes [1] shares a similar idea of training a face-swapping GAN with paired source video and the target video. Naruniec et al. [41] introduce the first approach that can swap facial identity at the megapixel resolution, again with the help of GAN. RSGAN [42] makes an early attempt of extracting facial identity and non-identity attributes separately for face swapping by the face and hair latent extractors. FSGAN [43] improves its previous work [44] by replacing the 3DMM fitting and re-rendering with a GAN-based face reenactment network. Although rapid progress has been made, there is still much room to improve in order to get realistic and high-fidelity results. StyleGAN [29] based works [59, 60, 62, 64] can generate high-fidelity swapped images, but they suffer from the limited representation space of StyleGAN which results in unsatisfied identity and non-identity preserving. FaceInpainter [34] explores face swap in heterogeneous domains based on StyleGAN [29] and 3DMM [4]. On the other hand, FaceShifter [35] introduces an adaptive framework that can integrate more faithful non-identity attributes from the target with the identity embedding of the source from a pre-trained face recognition model. This framework shows strong generalization and therefore many subsequent works [8, 17, 32, 58] follow a similar design. Simswap [8] designs Weak Feature Matching (weak FM) loss to preserve more detailed target attributes. However, it hurts the identity similarity between the source and the final swapped face. MobileFaceSwap [61] proposes a lightweight face swap model based on both Faceshifter [35] and Simswap [8]. HifiFace [58] improves identity similarity by adding 3DMM shape parameters into the facial identity feature, but the target facial identity still leaks into the resulting image. InfoSwap [17] tries to remove the target facial identity by masking out the identity-relevant features in all the convolution stages of the pre-trained face recognition model, but other target attributes such as background and expression can't be fully preserved, because, by definition, any face recognition model is trained to be invariant to all the non-identity attributes(i.e., pose, expression). In Smooth-swap [32], the face recognition model is further trained for face swapping by contrastive learning to smooth the identity representation space. Although the source identity preservation is improved, the preservation of non-identity attributes from the target also gets hurt. Xu et al. [6] propose a unified framework for face reenactment and face swapping, which shows that pixel-level and semantic attribute disentanglement can both help remove target facial identity and preserve non-identity attributes. In contrast, we delve into the U-net of encoder-decoder structure which is commonly used in previous works [17, 35], and unveil that skip-connection is one root cause for entanglement of the target identity and its non-identity attributes, we then propose novel regularization to encourage further disentanglement learning after removing the skip connections.

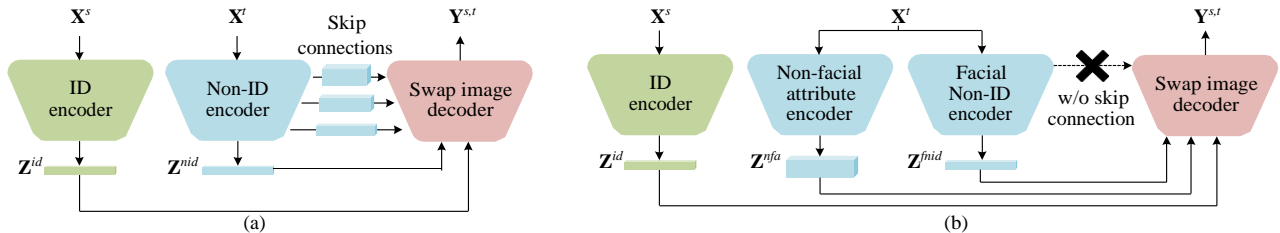

Figure 2. Framework comparison between (a) prior face swap methods w/ skip connections and (b) our method w/o skip connections.

**Face Disentanglement**. One mainstream face disentanglement idea is to leverage a geometry prior model, such as 3DMM [4], to disentangle identity, facial expression, head pose as well as lighting. Given a 2D face image, the goal is to estimate the 3DMM parameters either through direct optimization or deep learning models [11, 13, 14, 16, 63], then based upon the estimated parametric model, subsequent facial attributes editing may become easier. However, those prior models [36, 49] usually generalize poorly for wild images due to their limited capacity. Moreover, the fitting process also contains inevitable errors and ambiguities (i.e., between face identity and expression), limiting many applications that require precisely controllable editing. Therefore, a new line of work starts to learn disentangled facial attributes directly from 2D image-sets [9,23,31]. These generative approaches typically build latent representations for each specific facial attribute (i.e. pose, glass, hairstyle, elevation, etc.). Recent works [2, 5, 26, 45] focus more on identity and non-identity disentanglement. Specifically, Nitzen et al. [45] propose a latent space mapping network to map both the identity and attribute representations into StyleGAN latent code. Likewise, because of the limited information contained in StyleGAN latent code, this approach cannot preserve all the non-identity attributes, especially in hair and background. To better preserve the non-identity attributes, FICGAN [26] uses a much larger latent code to better restore expression and pose. However, the facial identity is not fully disentangled, meanwhile, the hair and background details are not totally preserved either.

**Skip Connection**. Skip connection is widely adopted in modern neural-net architectures, such as Highway [53], ResNet [22], DenseNet [24] and U-net [50] etc. It was initially designed to address the vanishing gradient issue when training very deep neural nets for classification tasks (i.e., image recognition [22]), but later extended to more broad applications including segmentation [19], optical flow prediction [46] and image synthesis [25, 35, 40]. Compared with short skip connections used in ResNet [22](via summation) and DenseNet [24](via concatenation), the symmetrical long skip connections used in U-net [50] directly copy fine-grained low-level details from the encoder to the decoder, results in more accurate and sharper dense predictions. However, we discover that even though skip connection benefits faster convergence, stable training, and finedgrained details preservation, it is one root cause for identity leaking and face attributes entanglement in the state-of-the-art face swap models [35].

## 3. Method

Our method aims to swap the facial identity of a source image $\mathbf{X}^s$ to a target image $\mathbf{X}^t$. The face-swapped result $\mathbf{Y}^{s,t}$ is a seamlessly blended version of the facial identity of $\mathbf{X}^s$ and all other identity-irrelevant(also called "non-identity") properties (*e.g.*, background, hair, pose, expression, and illumination) of $\mathbf{X}^t$. As discussed in Sec. 1, previous works commonly employ skip connections (see Fig. 2(a)), where shallow convolution embedding is fed to the swap image decoder. As shown in Fig. 2(b), instead of using skip connections, we propose a novel framework consisting of a Facial Non-ID (FNID) network and a Non-Facial attributes (NFA) network to perform face swapping. The skip-connection-free framework exhaustively preserves target non-identity information, while at the same time preventing the target facial identity from leaking into the swap image decoder. In the following, we first analyze the ID *vs.* Non-ID disentanglement of prior works through a performance calibration based on consolidated metrics. Then we dissect the effect of skip connections to motivate our new designs (Sec. 3.1). Finally, we introduce our proposed network structure and training strategies (Sec. 3.2) in detail.

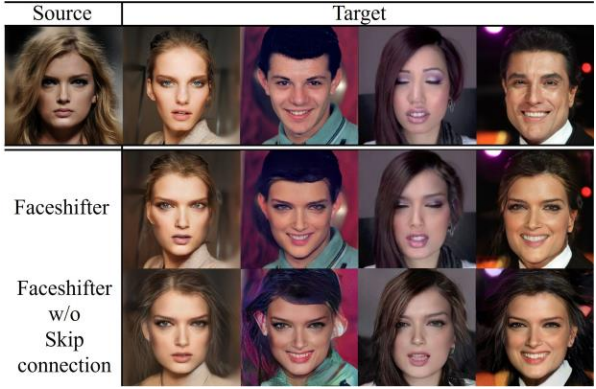

Figure 3. The effect of skip connection for Faceshifer [35]. Note that, removing the skip connection reduces the leakage of the target face identity but hurts the preservation of its non-identity attributes (*i.e.*, facial expression, hair details, *etc.*). Our work is designed to fix this issue.

| Methods | Use skip connections | Relative Performance | | |
|---|---|---|---|---|
| | | ID-ret | pose | expression |
| Faceshifter [35] | ✓ | / | / | / |
| Hififace [58] | ✓ | + 5% | - 7% | +0.3% |
| InfoSwap [17] | ✓ | +10% | -77% | -11% |
| Simswap [8] | ✗ | -19% | + 9% | +23% |
| Faceshifter* | ✗ | + 8% | -24% | -48% |

Table 1. Calibrated results among different prior works under the same evaluation method. * represents a version of Faceshifter after removing skip connections.

**ID *vs*. Non-ID disentanglement**. Although face swap has been vigorously studied for years, the evaluation methods for non-ID performance are distinct in previous reports [8, 17, 35, 58]. To consolidate the evaluations and perform fair comparisons, we employ off-the-shelf pose and expression estimations from the state-of-the-art 3D face reconstruction model [39] to serve as the non-ID metrics. And run their publicly released face swap results/models to obtain a quantitative ID/non-ID comparison, and finally, calibrate all results by treating Faceshifter [35] as a baseline. As shown in Tab. 1, after calibration, we discover that existing works still suffer from insufficient disentanglement between ID-relevant and ID-irrelevant representations, causing the improvement of a single metric always comes with the performance drop of others.

### 3.1. Dissection of Skip Connections

We noticed that in Faceshifter [35] and most follow-up works [17, 32, 58], skip connections play an important role in preserving non-identity attributes. Hence, we delve deep into this design. After diagnostic evaluations, we realized that because shallow convolutional features contain both ID and non-ID information, the skip connection that introduces shallow features to image decoding could be the "reason" behind such entanglement. By directly feeding the shallow convolutional features into the swap image decoder through skip connections, the detailed non-ID information (*e.g.*, background, hair, and expression) can be easily preserved. However, as shown in Fig. 3, we observe that the swapped facial identities of Faceshifter are noticeably affected by the target face shape, resulting in very low identity similarity to the source image. Alternatively, If we remove the skip connections in Faceshifter, surprisingly, the swapped facial identities are much closer to the source image, indicating less identity information from the target being leaked into the result. Nevertheless, it's also clear from Fig. 3 that such improvement comes with the cost of worse non-ID preservation (*e.g.*, hair). That is, image details out of the face region are inconsistent with target images. This phenomenon is also statistically confirmed in Tab. 1. As shown, removing the skip connection brings a better identity score but worse non-identity performance. Therefore, we conclude that the skip connection is at least one cause[1] that leads to the entangled representation.

### 3.2. Network Designs and Training Strategies

As we argued in Sec. 1, $\mathbf{Z}^{nid}$ alone (see Fig. 2 (a)) is not sufficient to preserve target non-ID properties without skip connections. In our method, we define two types of non-ID properties based on the relation with face ID, *i.e.*, face-relevant or face-irrelevant properties. Then, we design two encoders to extract the two-fold non-ID features, respectively. One is the **FNID** encoder that derives $\mathbf{Z}^{fnid}$ and the other is **NFA** encoder for generating $\mathbf{Z}^{nfa}$.

**Facial Non-ID (FNID) Encoder**. Facial Non-ID attributes (*e.g.*, expression, pose) are spatially coupled with the facial identity, thus we need deep convolution layers to remove the identity and preserve Non-ID attributes. Following this motivation, our FNID encoder only outputs the deepest representation $\mathbf{Z}^{fnid}$ without shallow features. We design training strategies tailored for FNID encoder. As shown in Fig. 4(a), to encourage the FNID encoder to extract facial Non-ID features and remove facial identities, we design two additional heads to regularize $\mathbf{Z}^{fnid}$, namely regularization head (RegHead) and adversarial head (AdvHead). RegHead maps $\mathbf{Z}^{fnid}$ into a vector that represents pose and expression information. In addition, a pre-trained 3DMM parameter predictor [14] is also used to predict pose and expression vectors $\mathbf{v}^{pose}$, $\mathbf{v}^{exp}$. Then, we use $l_2$ loss to regularize the mapped vector with the predicted pose and expression vectors:

$$L_r^{fnid} = \|\text{RegHead}(\mathbf{Z}^{fnid}) - \text{concat}(\mathbf{v}^{pose}, \mathbf{v}^{exp})\|_2^2 \quad (1)$$

---
[1]For those works that employ skip connections

AdvHead is to map $\mathbf{Z}^{fnid}$ into facial identity code, which is used to remove facial identities from $\mathbf{Z}^{fnid}$. In detail, AdvHead attempts to extract ID feature $\mathbf{Z}^{id}_{\mathbf{X}^t}$ of $\mathbf{X}^t$ from $\mathbf{Z}^{fnid}$ and we use an adversary loss to prevent $\mathbf{Z}^{fnid}$ from containing the ID information:

$$L_{adv}^{fnid} = \text{CSim}(\text{AdvHead}(\mathbf{Z}^{fnid}), \mathbf{Z}^{id}_{\mathbf{X}^t}) \quad (2)$$
$$L_{ah}^{fnid} = 1 - \text{CSim}(\text{AdvHead}(\mathbf{Z}^{fnid}), \mathbf{Z}^{id}_{\mathbf{X}^t}),$$

where $L_{ah}^{fnid}$ is only applied on the adversary head while $L_{adv}^{fnid}$ is applied on the FNID encoder. CSim denotes cosine similarity. The overall loss to train FNID can be expressed as:

$$L^{fnid} = L_r^{fnid} + \beta_{adv}^{fnid} L_{adv}^{fnid}. \quad (3)$$

**Non-Facial Attribute (NFA) Encoder**. Non-facial attributes are spatially separated from the facial regions, so they are weakly associated with the identity. Thus, a straightforward way to extract non-facial features without identity information is to mask out the facial regions of the target image $\mathbf{X}^t$ and use a masked-image encoder for feature extraction. However, this method would require a face mask predictor during the inference stage. To reduce inference complexity, instead of masking out face region from $\mathbf{X}^t$, we encourage our NFA encoder to process full $\mathbf{X}^t$ and extract similar representations to the masked-image encoder. In terms of network structure, NFA encoder has the same depth as FNID encoder but only the first few convolution layers have a down-sampling operation. Thus, the produced $\mathbf{Z}^{nfa}$ is a feature map with large spatial resolution. NFA encoder is trained in parallel with the masked-image encoder in the training stage (see *suppl. material* for details). We use a regularization loss on the $\mathbf{Z}^{nfa}$ and the $\hat{\mathbf{Z}}^{nfa}$ from masked-image encoder:

$$L^{nfa} = \|\mathbf{Z}^{nfa} - \hat{\mathbf{Z}}^{nfa}\|_2^2. \quad (4)$$

**Swap Image Decoder**. The swap image decoder takes $\mathbf{Z}^{fnid}$, $\mathbf{Z}^{nfa}$, $\mathbf{Z}^{id}$ as the input, and generate face-swapped image $\mathbf{Y}^{s,t}$. Three components are included, *i.e.*, FNID decoder, NFA decoder, and fusion network (shown in

$$L_{id} = \text{CSim}(\mathbf{Z}^{id}_{\mathbf{X}^s}, \mathbf{Z}^{id}_{\mathbf{Y}^{s,t}})$$

$$L^{rec} = \|\mathbf{X}^t - \mathbf{Y}^{t,t}\|_2^2$$

$$L_{attr} = \sum_{i=1}^{N^f} \|\mathbf{F}_{i,\mathbf{X}^t}^{fnid} - \mathbf{F}_{i,\mathbf{Y}^{s,t}}^{fnid}\|_2^2 + \sum_{j=1}^{N^a} \|\mathbf{F}_{j,\mathbf{X}^t}^{nfa} - \mathbf{F}_{i,\mathbf{Y}^{s,t}}^{nfa}\|_2^2,$$

$$L^{glb} = L_{id} + \beta^{rec} L_{rec} + \beta^{attr} L_{attr}. \quad (5)$$

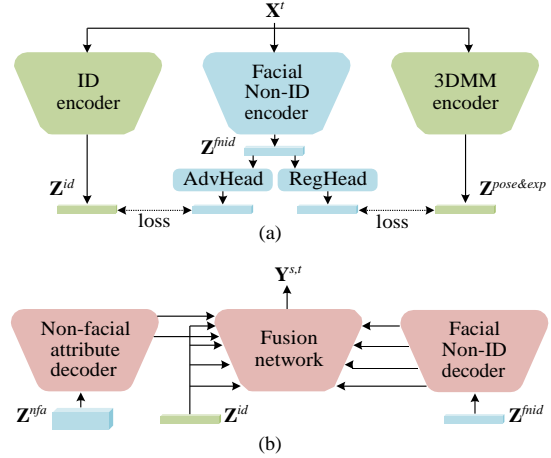

Figure 4. Details of network structure in our method. (a) Training heads for Facial Non-ID encoder. (b) Swap image decoder.

Fig.4(b)). FNID decoder decodes $\mathbf{Z}^{fnid}$ into $N^f$ feature maps $\{\mathbf{F}_i^{fnid}\}_{i=1}^{N^f}$ for fusion network. NFA decoder has similar structure that decodes $\mathbf{Z}^{nfa}$ into $N^a$ feature maps $\{\mathbf{F}_i^{nfa}\}_{i=1}^{N^a}$. Because Non-Facial attribute features are usually represented by shallow features, we set $N^a < N^f$. The fusion network integrates $\{\mathbf{F}^{fnid}\}$, $\{\mathbf{F}^{nfa}\}$ and $\mathbf{Z}^{id}$ for generating $\mathbf{Y}^{s,t}$. AAD layers [35] is employed as the fusion module. The number of AAD layers is $N^s$ to fully preserve non-ID attributes. In the first $N^f - N^a$ AAD layers, $\mathbf{F}_k^{fnid}(N^f > k > N^a)$ and $\mathbf{Z}^{id}$ are integrated. While in the last $N^a$ AAD layers, we concatenate $\mathbf{F}_k^{fnid}$ and $\mathbf{F}_k^{nfa}(k \leq N^a)$ and then integrate them with $\mathbf{Z}^{id}$. Following [35], we use the three global losses: ID loss, reconstruction loss, and attribute loss to train the whole network:

**ID Encoder**. Like Faceshifter [35], a pretrained face recognition model [12] is used as the ID encoder in our method. The normalized feature vector $\mathbf{z}^{id}$ extracted by the ID encoder is then fed to swap image decoder.

**Total Loss**. The overall training loss function is defined as:

$$L = L^{adv} + \beta^{glb} L^{glb} + \beta^{fnid} L^{fnid} + \beta^{nfa} L^{nfa}. \quad (6)$$

## 4. Results

### 4.1. Datasets and Metrics

**Training Data**. We use three face image datasets to train our model, *i.e.*, CelebA-HQ [28], FFHQ [29] and VGGFace

| Method | ID retrieval↑ | pose↓ | expression↓ |
|---|---|---|---|
| *evaluation w/ 3D face reconstruction model [39]* | | | |
| Deepfakes [1] | 81.96 | 0.375 | 7.270 |
| FSGAN [43] | 76.57 | 0.169 | 5.402 |
| Faceshifter [35] | 97.38 | 0.195 | 5.467 |
| Simswap [8] | 92.83 | <u>0.167</u> | **4.943** |
| Hififace [58] | 98.48 | 0.216 | 5.474 |
| InfoSwap [17] | <u>99.67</u> | 0.441 | 5.733 |
| MegaFS [64] | 90.83 | 0.465 | 5.615 |
| AFS [55] | 88.81 | 0.171 | 5.493 |
| UniFace [6] | 99.45 | 0.174 | 5.352 |
| Ours | **99.88** | **0.146** | <u>5.290</u> |
| *evaluation w/ the metrics of [62]* | | | |
| FSLSD [60] | 90.05 | 2.46 | – |
| StyleSwap [62] | 97.05 | 1.56 | 5.28 |
| StyleSwap w/ W+ [62] | 97.87 | **1.51** | 5.27 |
| Ours | **99.88** | **1.51** | **5.01** |

Table 2. Quantitative Comparison on FaceForensics++. The best and second best results are marked in **bold** and <u>underline</u>.

| Method | ID-CSim ↑ | ID-Consis ↑ |
|---|---|---|
| FSGAN [43] | 0.205 | 0.196 |
| Simswap [8] | 0.556 | 0.588 |
| InfoSwap [17] | 0.611 | 0.599 |
| MegaFS [64] | 0.466 | 0.543 |
| AFS [55] | 0.445 | 0.547 |
| Uniface [6] | <u>0.663</u> | <u>0.669</u> |
| Ours | **0.743** | **0.761** |

Table 3. Quantitative Comparison on CelebA-HQ test split. The best and second best results are marked in **bold** and <u>underline</u>.

[47]. As VGGFace contains many small and blur images, we only use 127.7K images from VGGFace with high facial resolution for training. Additionally, we select the first 100 images in CelebA-HQ for evaluation, so they are not involved in training. Following [35], all training images are cropped and aligned as 256 × 256 resolution to cover the entire face and a part of background regions. See *suppl. material* for more training details.

**Evaluation Data and Metrics**. FaceForensics++ (FF++) [51] is a standard face swap test dataset. On FF++, the overall performance of face swap methods is evaluated on three metrics: ID retrieval, pose error, and expression error. We follow the ID retrieval setting of most recent methods [1, 6, 8, 17, 35, 43, 58, 60, 62, 64], which uses a different face recognition model [56] as the ID vector extractor. However, the above-mentioned recent methods use different pose and expression predictors to evaluate pose and expression errors. For a fair comparison, we use the state-of-the-art 3D face reconstruction model [39] to predict pose and expression parameters and then employ Euclidean distance to define pose and expression errors. Also, we use pose [52] and expression [54] metrics reported by [62] to compare with recent methods [60, 62].

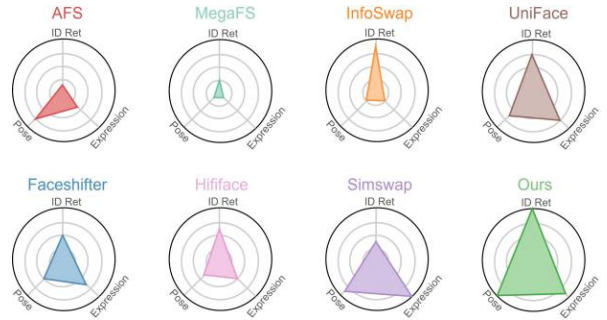

Figure 5. Comparison of face swap performance on FF++. Along each axis we plot the performance ranking of a metric, so a polygon with a larger area means better face swap performance.

In addition, we develop the CelebA-HQ test split with $N^c = 100$ samples to further analyze the ID swap ability from two-fold perspectives, *i.e.*, ID preservation from the source image and ID removal from the target image. Correspondingly, two additional metrics are designed, namely ID cosine-similarity (ID-CSim) for source-ID preservation and Swap ID consistency (ID-Consis ) for measuring the stability of one source image being swapped with varying target images. Specifically, for each image in the CelebA-HQ test split, we fix it as the source image and all other images serve as target images. ID-CSim is given by the cosine-similarities of ID vectors $\mathbf{Z}^{id}$ from the face recognition model [56] between the source image and corresponding swapped images, *i.e.*, ID-CSim = $\frac{1}{N^c(N^c-1)}\sum_{i=1}^{N^c}\sum_{j\neq i}\text{CSim}(\mathbf{Z}^{id}_{\mathbf{X}^i},\mathbf{Z}^{id}_{\mathbf{Y}^{i,j}})$. Then, we propose a novel ID-Consis metric to evaluate the ID consistency among the swapped images for one source image and different target images. If ID information is removed from target images, all the swapped images shall share a consistent facial ID. Thus, ID-Consis is given as the ID cosine-similarity for all swapped image pairs from the same source: ID-Consis = $\frac{\sum_{i=1}^{N^c}\sum_{j,k\in\{\mathbf{Y}^{i,\cdot}\}}\text{CSim}(\mathbf{Z}^{id}_j,\mathbf{Z}^{id}_k)}{N^c(N^c-1)(N^c-2)/2}$, where $\{\mathbf{Y}^{i,\cdot}\}$ is all swapped images with the source $\mathbf{X}^i$.

In terms of image quality, we compute the FID metric.

### 4.2. Comparison with Prior Arts

**ID *vs*. Non-ID Disentanglement**. Based on FF++, we first evaluate the overall performance of our method and recent face swap methods [1, 8, 17, 35, 43, 55, 58, 60, 64]. We use their original resolution in comparison because resolution has tiny effect on our ID and non-ID metrics. The works [60, 62] that release neither FF++ swapped images/videos nor pre-trained models are only compared with reported results. As shown in Tab. 2, our method achieves superior performances in terms of ID retrieval, pose error, and expression error. When compared with InfoSwap [17] which has strong ID swap ability, our results are still significantly better on all three metrics.

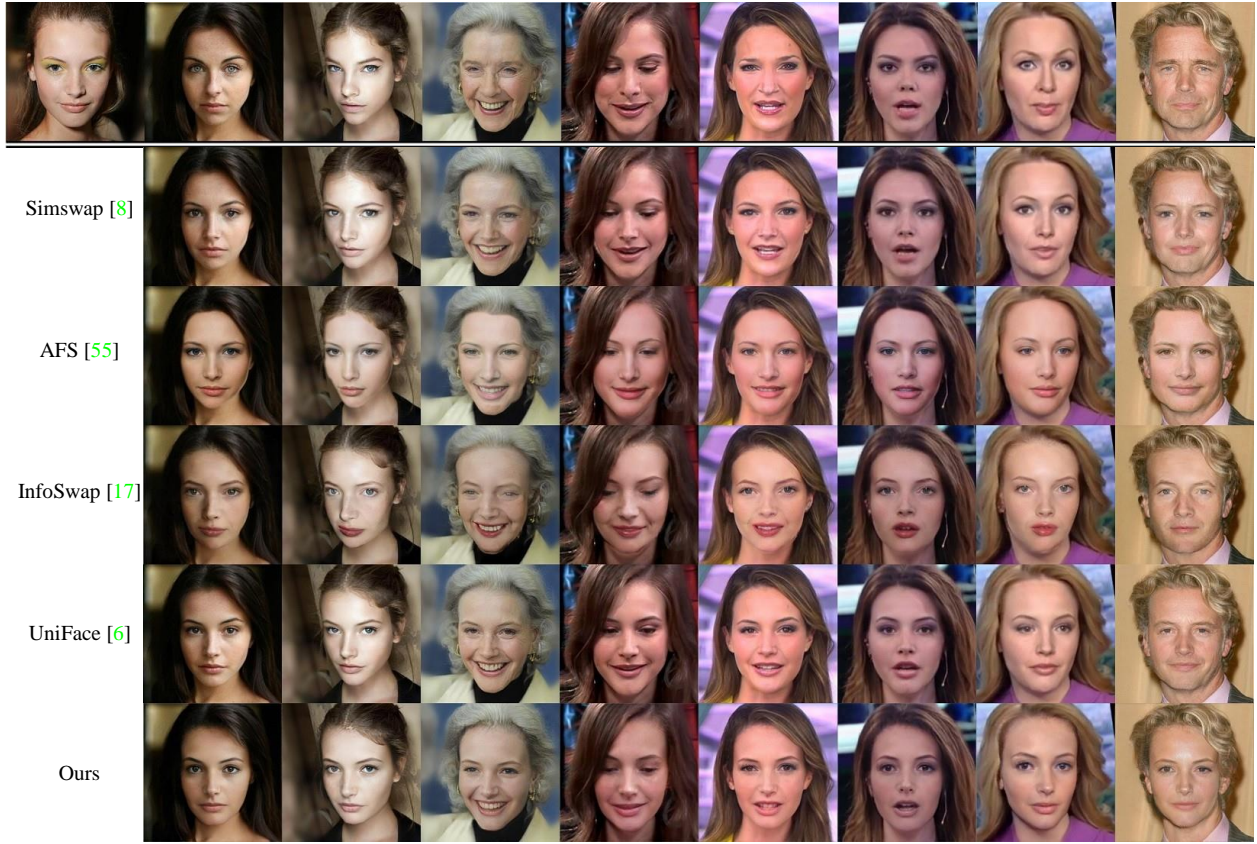

Figure 6. Comparison of the face-swapping results of various models. The top left corner is the source image, while the other images in the first row are target images. As shown, our results are noticeably better in terms of ID consistency across various target images.

|  | GAN based | | | StyleGAN based | |
|---|---|---|---|---|---|
|  | InfoSwap [17] | UniFace [6] | ours | FSLSD [60] | AFS [55] |
| FID↓ | 14.45 | 12.58 | 13.08 | 9.99 | 4.56 |

Table 4. Quantitative Comparison of image quality.

| Method | ID (%) | Non-ID (%) | Fidelity (%) |
|---|---|---|---|
| Faceshifter [35] | 17.6 | 22.0 | 25.4 |
| SimSwap [8] | 8.8 | 30.8 | 26.0 |
| InfoSwap [17] | 21.4 | 14.8 | 21.2 |
| Ours | **52.2** | **32.4** | **27.4** |

Table 5. User Study. Percentage of each method being selected.

Although Simswap [8] has the lowest expression error, we achieve better ID retrieval and lower pose error by a large margin. For a more intuitive comparison, we visualize the overall performance of the top 8 methods in Fig. 5 following [38], where our method achieves the best overall result with the largest pattern area. Meanwhile, Tab. 3 also compares the ID swap ability of our method with the recent methods that have publicly released models [8,17,43,55,64] on CelebA-HQ. As shown, our method achieves the best performance on both ID-CSim and ID-Consis, indicating that our method has better ID swap ability than compared methods. Considering that the current face recognition model is not perfect yet, our performances on ID-CSim and ID-Consis are quite convincing, as shown in Tab. 3. We argue those improvements are attributed to the better ID *vs.* non-ID disentanglement brought by our framework.

**Compare with StyleGAN-based methods**. In terms of image quality, though it is not our primary focus, the FID of our method is on par with the prior GAN-based methods, as shown in Tab. 4. Note that StyleGAN-based methods can produce significantly higher image quality than GAN-based methods due to the advantage of pre-trained StyleGAN [30] model. However, StyleGAN-based methods are poor in ID and non-ID preserving, which are more essential metrics for face swapping. Particularly, the SOTA StyleGAN-based method [62] reports much worse ID retrieval (2.01 drop) and expression error (0.26 increase) than ours (Tab. 2). This gap is mainly brought by the fact that the latent is not designed for ID and non-ID disentanglement and it is very difficult to preserve both ID and non-ID by latent swapping.

**Qualitative Evaluations**. In Fig. 1 and Fig. 6, we show examples of our results and the visual comparison with prior methods. Benefiting from the disentangled representation of our framework, our results have a noticeable advantage in terms of ID consistency across different target images

| Method | Skip connections | Facial Non-ID encoder | Regularization of $\mathbf{Z}^{fnid}$ | Non-facial attribute encoder | FaceForensics++ | | | | CelebA-HQ | |
|---|---|---|---|---|---|---|---|---|---|---|
| | | | | | ID-retrieval↑ | pose↓ | expression↓ | PSNR↑ | ID-CSim↑ | ID-Consis↑ |
| *Setting1* | ✓ | ✓ | ✗ | ✗ | 97.59 | 0.161 | 5.681 | **28.69** | 0.571 | 0.610 |
| *Setting2* | ✗ | ✓ | ✗ | ✗ | 99.41 | 0.175 | 5.801 | 24.74 | 0.726 | 0.735 |
| *Setting3* | ✗ | ✓ | ✓ | ✗ | 99.83 | **0.143** | 5.570 | 22.38 | 0.741 | 0.758 |
| Ours(full) | ✗ | ✓ | ✓ | ✓ | **99.88** | 0.146 | **5.290** | 28.33 | **0.743** | **0.761** |

Table 6. Ablation results. The best results are marked in **bold**.

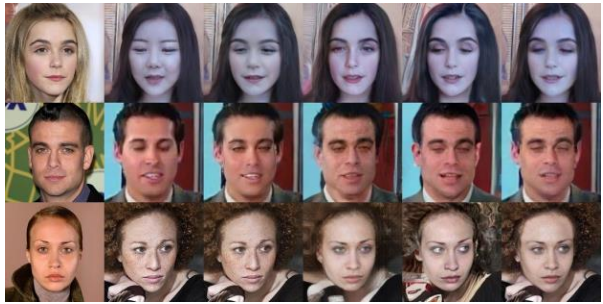

source    target *Setting1 Setting2 Setting3* Ours(full)

Figure 7. Visualization results of ablation models

when swapped with the same source. For rigorous qualitative evaluations, we further conduct user studies (see Tab. 5) on source-ID similarity, target non-ID preservation, and full image fidelity. The results indicate that our method significantly surpasses prior works on overall face swapping quality. Refer *suppl. material* for more details.

### 4.3. Ablation Studies

We perform ablation studies on FF++ and CelebA-HQ test split with all their evaluation metrics. To this end, three additional models are designed with different network components and training strategies, whose configurations and quantitative results are shown in Tab. 6. Besides, a few randomly selected rich expression results are shown in Fig 7.

**Effects of Skip Connections**. With skip connections, target facial ID could leak into the decoder through shallow features, leading to poor source-ID similarity. As shown in Tab. 6, compared to *Setting1*, skip-connection free of *Setting2* achieves significant improvements in all source-ID-related metrics. In contrast, skip connections also have a noticeable effect on non-ID attribute preservation from target images, especially for background details. As expected, *Setting2* is worse than *Setting1* in terms of PSNR, pose, and expression errors. Therefore, skip connections could lead to a seesaw-like trade-off by leading to the aforementioned compromises, which is not conducive to overall face-swapping performance.

**Effects of $\mathbf{Z}^{fnid}$ Regularization**. $\mathbf{Z}^{fnid}$ regularization strategy can simultaneously enhance the non-ID preservation and ID removal from target images. From *Setting2* to *Setting3*, as shown in Tab. 6, the pose and expression errors are both reduced thanks to this regularization, which is also confirmed by visual examples in Fig 7 (*i.e.*, eye closure&gaze is preserved in *Setting3*). Meanwhile, source-ID-related metrics are also elevated because target-ID are successfully removed by our adversarial head. Nevertheless, because of the limited capacity of $\mathbf{Z}^{fnid}$, the quality of facial non-ID and non-facial attributes can hardly be improved simultaneously, resulting in a poor PSNR value.

**Effects of NFA Modules**. The NFA module can model non-facial regions so as to improve the performances on both non-facial and non-ID information preservation. Referring to our full method in Tab. 6, PSNR is significantly elevated by NFA module and the value is on par with that of *Setting1*. In the meantime, The expression error is also noticeably reduced thanks to the NFA module. The reason behind this is that the non-facial attributes are represented by $\mathbf{Z}^{nfa}$, so $\mathbf{Z}^{fnid}$ can now dedicate modeling facial non-ID features. As the expression is a highly abstract non-ID property, a high-capacity $\mathbf{Z}^{fnid}$ is beneficial in modeling the extensive-expression variation. In contrast, the pose error remains nearly unchanged with NFA modules because the pose is a weak semantic property that is not sensitive to the capacity of $\mathbf{Z}^{fnid}$. As shown in Fig. 7, a significant improvement in expression and background detail preservation is introduced by our full method.

## 5. Conclusions

In this work, we unveil that the skip connection that was widely used in prior works is one root cause for poor disentanglement between ID and non-ID representation. We proposed a new framework to address this issue from both network structure and regularization loss perspectives. The experimental results confirm both our hypothesis and the effectiveness of our method.

**Limitation and future works**. StyleGAN-based face swap [58, 62] methods are superior in terms of rendering quality (*i.e.*, FID), although still suffer the disentanglement challenges. Combining our framework with StyleGAN would be interesting future work.

# Reinforced Disentanglement for Face Swapping without Skip Connection

## –Supplementary Material

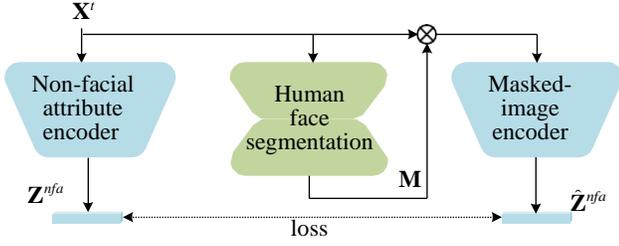

Figure 1. Details of NFA structure in training stage.

## A. Training the NFA encoder

To encourage the NFA encoder to extract the non-facial-region features, a masked-image encoder $\mathcal{M}$ is trained simultaneously with the NFA encoder during the training stage. As shown in Fig.1, a pre-trained human face segmentation model is used to predict face mask $\mathbf{M}$, and then $\hat{\mathbf{Z}}^{nfa}$ is generated with $\mathbf{M}$ and $\mathbf{X}^t$:

$$\hat{\mathbf{Z}}^{nfa} = \mathcal{M}((1-\mathbf{M}) \cdot \mathbf{X}^t), \quad (1)$$

where $(1-\mathbf{M})$ is the non-face mask.

In this way, facial-region features are excluded in $\hat{\mathbf{Z}}^{nfa}$. Thus we can remove facial region features from $\mathbf{Z}^{nfa}$ by using a regularization loss between $\mathbf{Z}^{nfa}$ and $\hat{\mathbf{Z}}^{nfa}$. The overall loss also encourages $\mathbf{Z}^{nfa}$ to contain non-facial region features as much as possible because $\mathbf{Z}^{nfa}$ is spatially larger than $\mathbf{Z}^{fnid}$ and has a stronger capacity of detail preservation.

## B. Training And Evaluation Details

**Architecture Details**. The detailed network structures of FNID and NFA modules are shown in Tab.1 The FNID encoder contains 7 down-sampling convolutional layers, while the NFA encoder contains 4 down-sampling convolutional layers and 3 ResBlocks. AdvHead is a 3-layer MLP that outputs a 512-dim vector with hidden layer size of 1024. The AdvHead is designed to be stronger to erase ID information from $\mathbf{Z}^{fnid}$. Besides, both ID encoder and AdvHead are the pre-trained ArcFace [4] face recognition model, and the pre-trained 3DMM predictor from [5] is used to form regularization loss $L_r^{fnid}$. RegHead is an FC layer to output a 67-dim vector with 3-dim $\mathbf{v}^{pose}$ and 64-dim $\mathbf{v}^{exp}$. In the last several layers of our Fusion network, the AAD have three inputs: $\mathbf{F}_k^{fnid}$, $\mathbf{F}_k^{nfa}$ and $\mathbf{Z}^{id}$. $\mathbf{F}_k^{fnid}$ and $\mathbf{F}_k^{nfa}$ are firstly concatenated to predict $\beta$ and $\gamma$, i.e., $\beta, \gamma = \text{Conv}([\mathbf{F}_k^{fnid}, \mathbf{F}_k^{nfa}])$. The usage of $\mathbf{Z}^{id}$ is the same as that in FaceShifter.

|  | FNID | NFA |
|---|---|---|
| Encoder | Conv(c= 32, s=2) | Conv(c= 32, s=2) |
|  | Conv(c= 64, s=2) | Conv(c= 64, s=2) |
|  | Conv(c= 128, s=2) | Conv(c=128, s=2) |
|  | Conv(c= 256, s=2) | Conv(c=256, s=2) |
|  | Conv(c= 512, s=2) | ResBlk(c=512, s=1) |
|  | Conv(c=1024, s=2) | ResBlk(c=512, s=1) |
|  | Conv(c=1024, s=2) | ResBlk(c=512, s=1) |
| Decoder | TConv(c=1024, s=2) | TConv(c=256, s=2) |
|  | TConv(c= 512, s=2) | TConv(c=128, s=2) |
|  | TConv(c= 256, s=2) | TConv(c= 64, s=2) |
|  | TConv(c= 128, s=2) | TConv(c= 32, s=2) |
|  | TConv(c= 64, s=2) |  |
|  | TConv(c= 32, s=2) |  |

Table 1. The FNID and NFA module details. Conv is the standard convlutional layer. TConv is the transposed convlutional layer. ResBlk is the residual convlutional block [7]. "c" is the number of output channels, and "s" denotes the up/down-sampling scales.

**More details of the training losses**. The adversary loss $L_{adv}$ is a Hinge GAN loss from a multi-scale (256, 128, 64) discriminator. The term $L_{ah}^{fnid}$ is to train the AdvHead. When it is used, the whole FNID encoder is fixed except the AdvHead. In contrast, $L^{fnid}$ is to train FNID encoder with RegHead, thus $L_{ah}^{fnid}$ does not contribute to the $L^{fnid}$.

**Hyper-parameters**. For balance and stable training, we set $\beta_{adv}^{fnid} = 0.1$ in $L^{fnid}$; $\beta^{rec} = 0.2$ and $\beta^{attr} = 0.5$ in $L^{glb}$; $\beta^{glb} = 5$, $\beta^{fnid} = 2$, $\beta^{nfa} = 100$ in the overall loss.

**Details of Comparison with StyleGAN-based Methods**. Recent SOTA StyleGAN-based face swap methods [13, 14] have not released their inference models or face swap results on FF++, thus we evaluate our method using the same pose and expression metrics of [14] for fair comparison. In terms of FID score, we only compare with [13] because the work [14] did not report the FID evaluation details.

**User Study Conduction Details**. We conduct a user study to evaluate the face swap performance from three perspec-tives: source ID similarity, target non-ID preservation, and image quality. To this end, we randomly select 100 source-target pairs from the FF++ [10] testset. The swap results are from faceshifter [8], Simswap [3], InfoSwap [6], and our method. Then, participants are asked to select: (i) the one with the best source-image ID similarity; (ii) the one with the best target-image similarity of the pose and facial expression; and (iii) the one that looks the most like the real photo.

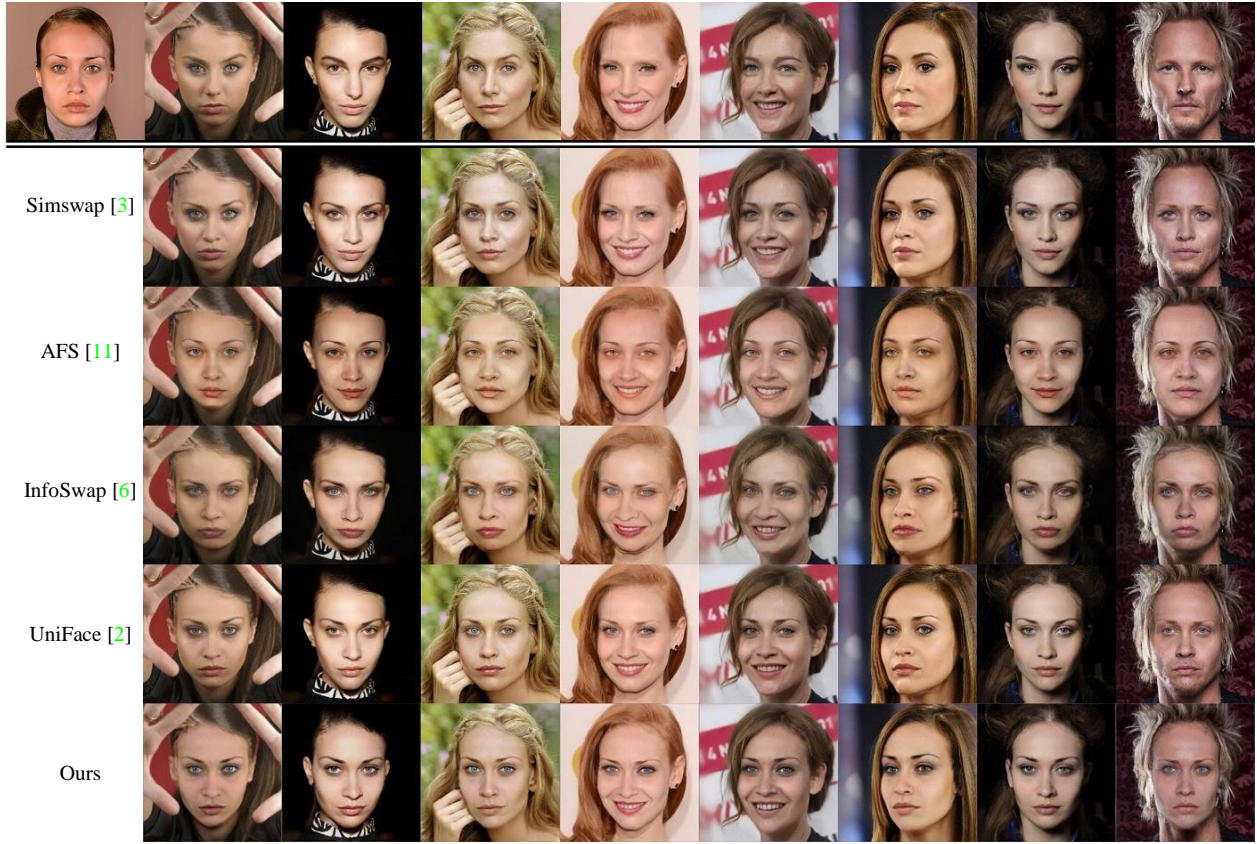

Figure 2. Comparison of ID consistency. The top-left corner is the source image, while the other images in the first row are target images.

## C. More Qualitative Results

**Comparison of ID Consistency**. We believe ID consistency is important in many applications (*e.g.*, virtual human creation, film-making), and swap identity should be consistent across various contents. Therefore we show additional results in Fig.2, where our method is superior in ID consistency.

**Comparison with Prior Arts in FF++**. To further visually compare our method with prior methods, we randomly collect source-target pairs in FF++. Referring to Fig. 3, we can see that our swap results are better than that of other methods in terms of source-ID similarity and target-non-ID preservation, indicating our method has advantages in the disentanglement representation.

**Comparing with Simswap**. Although Simswap [3] achieves slightly lower expression error than our method in quantitative comparison, its performance on source-ID similarity lags considerably behind our method. In Figs. 2 and 3, Simswap has minor visual advantages in expression preservation, but its swap results are not similar to the source ID. For example, referring to the 2nd-8th columns in Fig.2 and the all results in Fig.3, the swapped ID of Simswap are quite close to the target images. In contrast, our results are overall superior when compared to Simswap.

**Cross-Age/Gender/Hairstyle Face Swap Results**. As shown in Fig.4, our method can produce impressive face swap results for difficult cross-age/gender cases.

As for cross-hairstyle face swap, there are two situations as shown in Fig.5: (1) Target has bangs while source has no bang. Our method can handle this situation because our NFA encoder can detect the bangs as non-facial attributes and our decoder will preserve them. (2) Source has bangs while target has no bang. Our method cannot handle this situation because the pretrained ID encoder regards bangs as a part of facial ID.

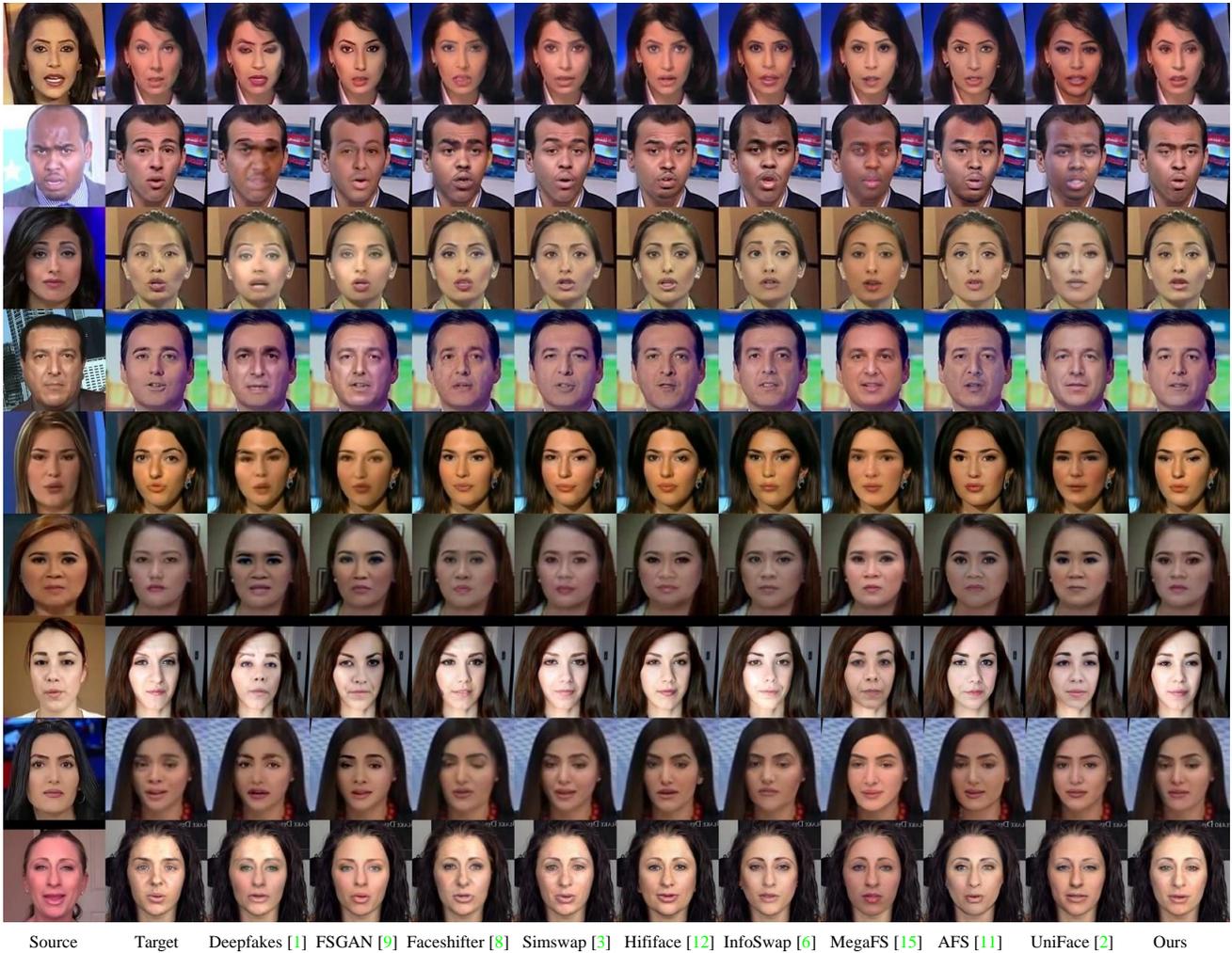

Figure 3. Comparison of face-swapping results on FF++.

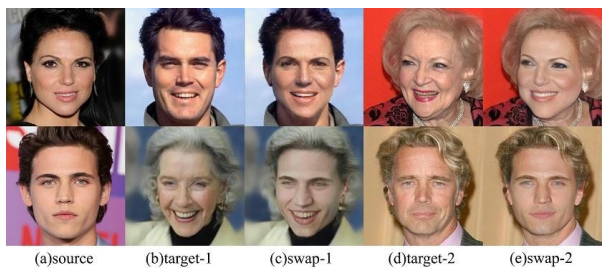

Figure 4. Cross gender and age results by our method.

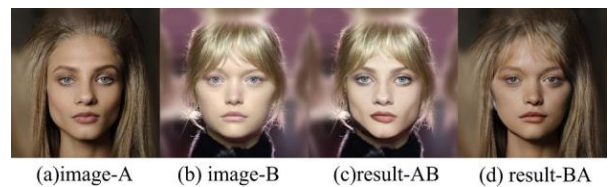

Figure 5. Cross hairstyle results by our method. Result-AB is produced by using image A as the source image and image B as the target image, result-BA is produced by using image B as the source image and image A as the target image.

## D. Discussion on Face shape swap

Face shape is a essential part of face ID. In our method design, the learned masks in AAD ResBLK affect significantly to face shape swap. Fig. 6(d) shows the learned masks in last four AAD ResBLK when face shape changes. From the masks, we find out the 7th AAD ResBLK plays important role in face shape swap. Fig. 6 (e) shows the details of mask of the 7th AAD ResBLK. The GREEN and BLUE lines roughly represent the swap and target face shape. The inner region of the BLUE line are very dark and the outer region of GREEN line is very light, which means those regions are generated mostly according to one input feature (either ID or non-ID). The region between the BLUE and GREEN line is lighter than the facial region yet darker than the background, which means this region should be generated according to both ID and non-ID features. From the qualitative results, our method is with better swapped face shapes than other methods.

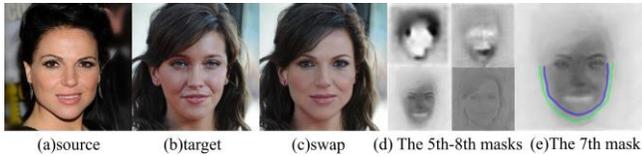

Figure 6. Visualization of the masks when face shape changes.

## E. Ethical Consideration

The goal of this paper is to study high-quality face swaps. It does not intend to manipulate existing images or to create misleading or deceptive content. However, the method, like all other related AI image generation techniques, could still potentially be misused for impersonating humans. We condemn any behavior to create such harmful content. Currently, the synthesized portraits by our method contain certain visual artifacts that can be identified by humans and some deepfake detection algorithms. We encourage to apply this method for learning more advanced forgery detection approaches to avoid potential misusage.